\pdfoutput=1
\documentclass{article}

\usepackage[preprint]{neurips_2026}

\usepackage[utf8]{inputenc}
\usepackage[T1]{fontenc}
\usepackage{hyperref}
\usepackage{url}
\usepackage{booktabs}
\usepackage{amsfonts}
\usepackage{nicefrac}
\usepackage{microtype}
\usepackage{xcolor}

\usepackage{placeins}
\usepackage{amsmath}
\usepackage{graphicx}
\usepackage{cleveref}

\definecolor{cornflowerblue}{rgb}{0.39, 0.58, 0.93}
\hypersetup{
    colorlinks=true,
    linkcolor=cornflowerblue,
    filecolor=magenta,
    urlcolor=teal,
    citecolor=cornflowerblue,
    pdftitle={Adaptive-Depth Is a Finishing Stage},
    }

\title{Adaptive-Depth Is a Finishing Stage}

\author{Andrei Piterbarg$^{1}$\thanks{Correspondence to: \texttt{andreip@engineering.upenn.edu}.} , Sean McLeish$^{2}$, Dhruv Pai, Benjamin Keigwin\\
$^{1}$ University of Pennsylvania,
$^{2}$ University of Maryland,
}

\begin{document}

\maketitle

\begin{abstract}
Looped transformers reuse the same layer multiple times, allowing inference compute to be adjusted without changing the model.
Recent adaptive-depth methods train the model across multiple loop depths from the beginning of training.
We introduce \textit{fixed-first training}, which varies model depth only during a short final phase of training.
At 200M parameters, training at fixed depth 2 for the first 67\% of the run reduces modeled compute by 45\% while keeping mean negative log-likelihood (NLL) across inference depths within 0.5\% of training across depths throughout.
At 1B parameters, training initially at depth 4 reduces modeled compute by 31\% to 37\% and wall-clock time by 32\% to 39\%, while keeping terminal NLL within 0.8\% and downstream differences within the observed seed variation.
These results show that training over the full range of depths can be reserved for a short finishing phase with only small changes in performance.
\end{abstract}

\section{Introduction}
The depth of a standard decoder language model is the number of unique layers used sequentially per token.
A looped transformer instead applies the same layer, with the same weights, several times.
Our models use the middle-loop architecture of \citet{saunshi2025reasoning,geiping2025scaling}, in which an untied prelude and coda surround a recurrent block which is repeated to increase model depth.
We use \texttt{K} to denote the number of times this block is applied, and we refer to the depth chosen at inference as the serving depth.
One of the main benefits of looped transformers is that one model can be adaptive: it can give useful answers at small serving depths and improve as we increase the depth and spend more compute.
Training at a single depth can produce a depth-specialized model, so existing adaptive models train across multiple depths to perform well across the serving range.

Recent adaptive-depth methods expose the model to multiple depths throughout training through different recipes.
\citet{geiping2025scaling} sample one recurrent depth from a distribution for each training step.
\citet{mcleish2025teaching} reduce this cost with a curriculum that progressively increases the mean of a Poisson-lognormal depth distribution over training.
Other approaches supervise multiple trajectories or exits~\citep{jeddi2026loopformer,zhu2025scaling}.
We compare fixed-first with two sampled-depth baselines: uniform-depth training, also used by \citet{hegazy2026gated}, and the progressive Poisson-lognormal curriculum of \citet{mcleish2025teaching}.
Uniform-depth training samples \texttt{K} uniformly from \texttt{\{1,...,16\}} on every update.
This exposes the model to the full serving range throughout training, but it also requires expensive high-depth updates from the beginning.

\looseness -1
Fixed-first instead makes adaptive-depth training a finishing stage rather than a cost paid throughout the run: it holds the training depth constant for most of training and only introduces depth variation at the end.
Specifically, fixed-first trains at an initial depth \(K_0\) for the first 67\% or 80\% of training, then switches to uniform \texttt{K=1,...,16} training.
Because recurrent compute grows with \texttt{K}, using a smaller \(K_0\) during the fixed phase reduces the number of layer applications and therefore the modeled training Floating Point Operations (FLOPs).
Fixed-first changes only the recurrent depths used during training, not the architecture, parameter count, or token budget; \Cref{fig:schedule} shows an 80\% fixed-first schedule.

Our contributions are:
\begin{enumerate}
    \item We motivate fixed-first training using a controlled factual-recall experiment that shows why the method has two stages: fixed-depth training learns quickly but specializes near its training depth, while uniform training retains recall over a broader range of depths but learns more slowly.
    \item At 200M, varying \(K_0\) reveals that the initial fixed depth controls both compute savings and the serving depths the final model favors. For example, using \(K_0=2\) for 67\% of training saves 45\% of modeled compute while keeping mean serving-depth NLL within 0.5\% of uniform training.
    \item At 1B, fixed-first training reduces modeled training compute by 31\% to 37\% while keeping terminal NLL within 0.8\% of uniform training and downstream evaluations.
    \item Finally, we use an update diagnostic to examine why the finishing phase can be short.
\end{enumerate}

\section{Factual Recall Motivates Fixed-First Training}

To begin, we use a controlled synthetic memorization task to test how models learn new facts.
The task isolates factual storage: training and evaluation use the same people and facts, while varying how those facts are presented.
This task is based on the bioS datasets introduced by \citet{allen2023physics}; additional task details and a fixed-depth specialization sweep are provided in \Cref{app:synthetic-task}.

Our dataset is made up of fictional people, each with six attributes: birth city, birth day, birth month, university, major, and employer.
Together, the six attributes contain \(33.39\) bits of information per person, based on the total number of possible combinations.
We test each fact using five fixed templates, and a fact is only considered recalled if the model gives the exact answer on at least four of the templates.
To calculate the recalled bits per person, we note that each recalled attribute contributes $\log_2(\text{number of possibilities})$ bits, which we then sum across all the attributes.

\begin{figure}[!t]
  \centering
  \includegraphics[width=0.85\linewidth]{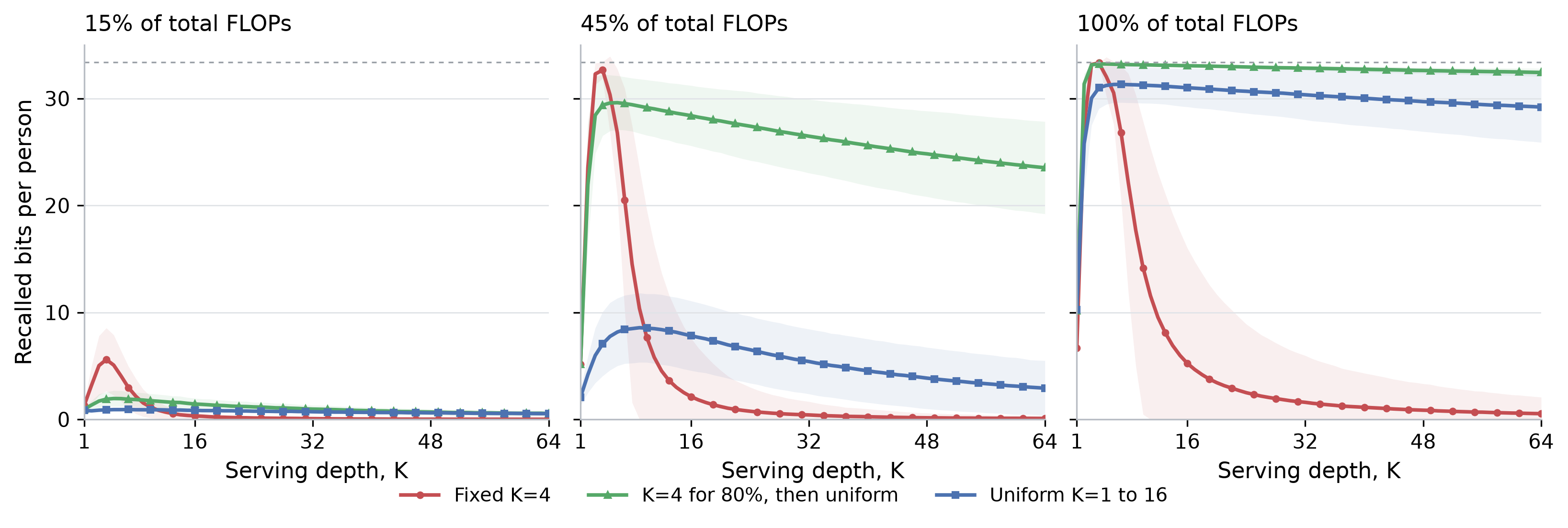}
\caption{Recalled bits per person across serving depths for fixed \texttt{K=4}, an 80\% fixed-first schedule, and uniform \texttt{K=1,...,16} training at three matched training budgets (15\%, 45\%, and 100\% of the full FLOP budget). Panels compare complete runs: we train a new model of each schedule from scratch at each budget, so the fixed-first models have switched to uniform training in each panel. Lines show four-seed means, bands show 95\% confidence intervals, dashed line marks the task ceiling.}
  \label{fig:memorization-fixed80}
\end{figure}

Using this substrate, we compare fixed \texttt{K=4}, fixed $K_0=4$ for 80\% of the updates followed by uniform \texttt{K=1,...,16}, and uniform \texttt{K=1,...,16} throughout all of training in \Cref{fig:memorization-fixed80}.
The three schedules are compared at matched modeled training FLOPs.
Although training depths never exceed \texttt{K=16}, we evaluate recall through \texttt{K=64} to test how each schedule behaves beyond the training range.

In the middle panel of \Cref{fig:memorization-fixed80}, with 45\% of the training budget, fixed \texttt{K=4} has learned quickly near its training depth but its recall collapses at larger serving depths.
Uniform training is more stable across depths but has learned much less. Because the runs are matched on FLOPs rather than updates, and a \texttt{K=4} update costs less than an average uniform update, the uniform run takes fewer optimizer steps, which may account for part of this gap.
With the full budget, fixed-first training remains close to the task ceiling across serving depths through \texttt{K=64}. Uniform training also develops broad recall, but declines beyond the \texttt{K=1,...,16} range used during training. We read these curves only as differences in peak recall and in how quickly recall falls off with serving depth; they do not show whether the slower decline of fixed-first reflects a converged, depth-independent computation.
\Cref{fig:fixed-depth-specialization} shows models trained at \texttt{K=1,2,4,8} also perform best near their respective training depths.

This is a comparison of factual-recall learning speed, not final storage capacity.
\citet{zhu2025scaling} show that iso-parameter models with different recurrent depths reach essentially the same knowledge capacity.
Here, fixed-first training achieves useful recall sooner and retains more of that recall beyond the depth range used during training.

\section{Language-Model Results}
Next, we test whether the gains of fixed-first training carry over to language modeling.
We train 200M-parameter looped transformers from scratch on UltraData-Math-L3~\citep{wang2026data} for 4,800 updates, or 2.52B tokens, and evaluate them at serving depths \texttt{K=1,...,16}.
We vary $K_0\in\{1,2,4\}$, switch after 67\% or 80\% of training, and use either a uniform or Poisson-lognormal finish; uniform training and the progressive curriculum of \citet{mcleish2025teaching} are references.

\begin{figure}[!t]
  \centering
  \includegraphics[width=0.85\linewidth]{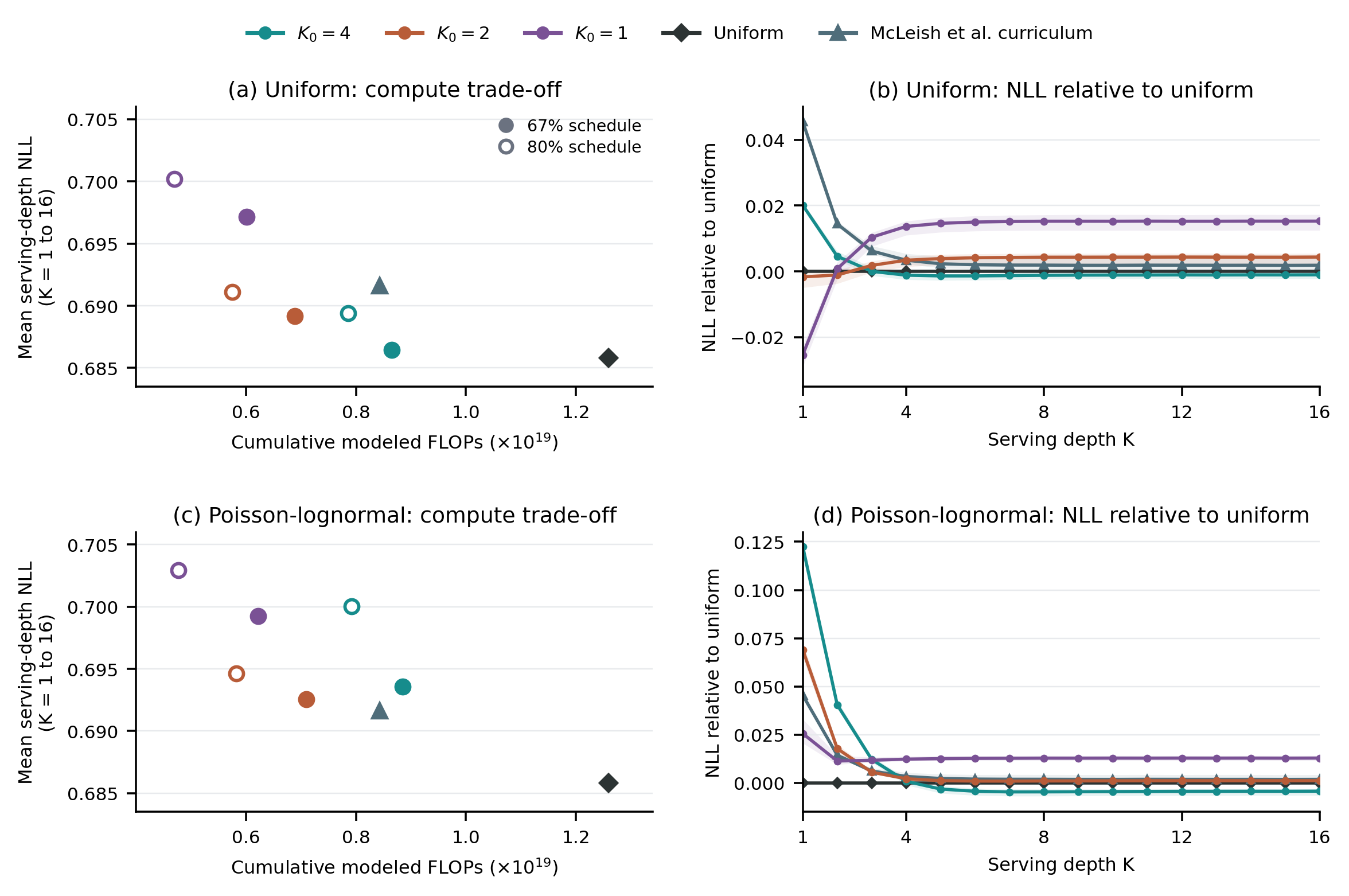}
  \caption{Results at 200M. Rows show the finishing distribution. Left: mean serving-depth NLL against modeled training FLOPs; filled and open markers denote 67\% and 80\% fixed phases. Right: serving-depth NLL minus uniform-training NLL for the 67\% schedules; positive values are worse than uniform. Lines and bands show the mean and range across three paired seeds.}
  \label{fig:anchor-depth}
\end{figure}

With a uniform finish, $K_0=2$ cuts modeled compute by 45.3\% at the 67\% switch while keeping mean serving NLL within 0.5\% of uniform training; at the 80\% switch it saves 54.3\% while remaining within 0.8\% (\Cref{fig:anchor-depth}).
Lowering $K_0$ shifts quality toward cheaper serving depths: at the 67\% switch, $K_0=1$ has the best \texttt{K=1} NLL but the ordering reverses from \texttt{K=3} onward.
Uniform finishing gives lower mean NLL than Poisson-lognormal finishing for all six fixed-first schedules.
Compared with the \citet{mcleish2025teaching} curriculum, the $K_0=2$ uniform-finish schedules reach lower mean NLL with 18.2\% less modeled compute at the 67\% switch and 31.7\% less at the 80\% switch. The \(K_0=4\), 67\% schedule nearly matches full uniform training in mean serving-depth NLL (0.68643 versus 0.68582) with 31.3\% less modeled compute; the compute-matched uniform run reaches 0.72089.

\begin{figure}[t!]
  \centering
  \includegraphics[width=0.75\linewidth]{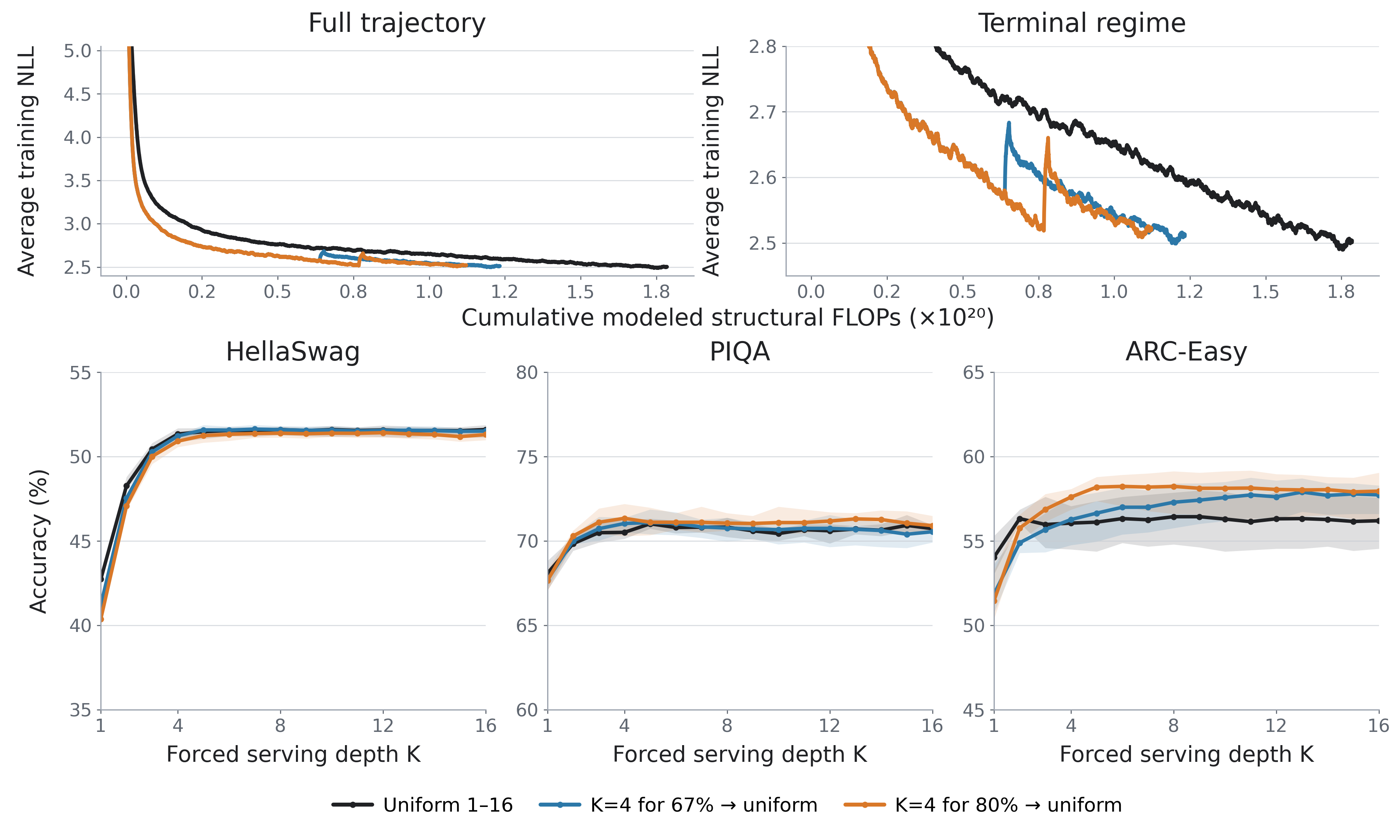}
  \caption{Results at 1B. Top: full and terminal training NLL trajectories. The loss spike is where we switch from fixed to adaptive-depth training. Bottom: HellaSwag~\citep{zellers2019hellaswag}, PIQA~\citep{bisk2020piqa}, and ARC-Easy~\citep{clark2018think} accuracy across serving depths. Lines are three-seed means and bands span seeds.}
  \label{fig:one-billion-results}
\end{figure}

Next, we scale model size to a 1B looped transformer training on Ultra-FineWeb~\citep{wang2025ultra}.
Each model trains for 20,000 updates (10.49B tokens); we compare uniform training with $K_0=4$ fixed-first schedules that switch after 67\% or 80\% of training.

At 1B, the 67\% and 80\% schedules cut training FLOPs by 30.8\% and 36.9\% and wall-clock time by 32.1\% and 38.5\%, respectively.
Terminal NLL remains within 0.8\% of uniform training, and the three evaluations in \Cref{fig:one-billion-results} remain within the observed seed variation; exact results are in Appendix~\ref{app:language-model-results}.

\begin{figure}[!t]
  \centering
  \includegraphics[width=0.65\linewidth]{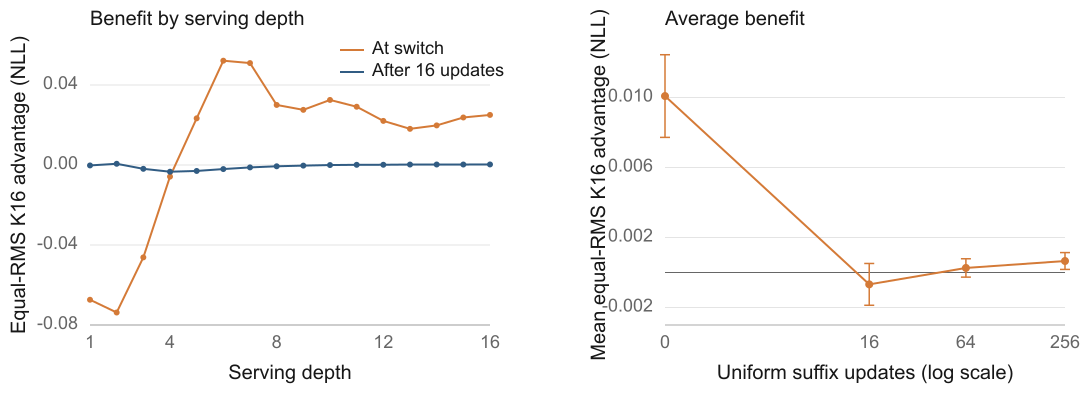}
  \caption{We define the benefit as the NLL after the \texttt{K=4} update minus the NLL after the \texttt{K=16} update; positive values favor \texttt{K=16}. Left: the benefit across serving depths at the start of the uniform phase and after 16 uniform updates. Right: the benefit averaged across serving depths after 0, 16, 64, and 256 uniform updates; error bars show 95\% confidence intervals.}
  \label{fig:mechanism}
\end{figure}

\section{Why the Finishing Stage Can Be Short}

To understand why fixed-first training matches uniform-depth training, we compare single updates at \texttt{K=4} and \texttt{K=16} from the same checkpoint, batch, and optimizer state.
We isolate the part of each update caused by the batch, match their RMS, and evaluate them across the full serving range.

At the start of the uniform phase, the deeper update improves the model at larger serving depths, while the shallower update is more useful near \texttt{K=4} (\Cref{fig:mechanism}).
After 16 uniform updates, the depth-averaged advantage of the \texttt{K=16} update is no longer distinguishable from zero.
This suggests that the finishing phase quickly corrects depth specialization once a strong low-depth computation is present, but does not prove that deeper updates are unnecessary over the full trajectory.

\section{Conclusion}
\looseness -1
Across 200M and 1B models, fixed-first training cuts 30\% to 40\% of modeled training compute and 26\% to 39\% of wall-clock time while keeping NLL within 1\% and matching downstream performance.
Our diagnostic suggests that deep updates remain important for flexibility, but their relative advantage becomes small soon after the uniform phase begins.

\clearpage
\bibliography{refs}

\begin{thebibliography}{13}
\providecommand{\natexlab}[1]{#1}
\providecommand{\url}[1]{\texttt{#1}}
\expandafter\ifx\csname urlstyle\endcsname\relax
  \providecommand{\doi}[1]{doi: #1}\else
  \providecommand{\doi}{doi: \begingroup \urlstyle{rm}\Url}\fi

\bibitem[Allen-Zhu and Li(2023)]{allen2023physics}
Zeyuan Allen-Zhu and Yuanzhi Li.
\newblock Physics of language models: Part 3.1, knowledge storage and
  extraction.
\newblock \emph{arXiv preprint arXiv:2309.14316}, 2023.

\bibitem[Bisk et~al.(2020)Bisk, Zellers, Gao, Choi, et~al.]{bisk2020piqa}
Yonatan Bisk, Rowan Zellers, Jianfeng Gao, Yejin Choi, et~al.
\newblock Piqa: Reasoning about physical commonsense in natural language.
\newblock In \emph{Proceedings of the AAAI conference on artificial
  intelligence}, volume~34, pages 7432--7439, 2020.

\bibitem[Clark et~al.(2018)Clark, Cowhey, Etzioni, Khot, Sabharwal, Schoenick,
  and Tafjord]{clark2018think}
Peter Clark, Isaac Cowhey, Oren Etzioni, Tushar Khot, Ashish Sabharwal, Carissa
  Schoenick, and Oyvind Tafjord.
\newblock Think you have solved question answering? try arc, the ai2 reasoning
  challenge.
\newblock \emph{arXiv preprint arXiv:1803.05457}, 2018.

\bibitem[Geiping et~al.(2025)Geiping, McLeish, Jain, Kirchenbauer, Singh,
  Bartoldson, Kailkhura, Bhatele, and Goldstein]{geiping2025scaling}
Jonas Geiping, Sean McLeish, Neel Jain, John Kirchenbauer, Siddharth Singh,
  Brian~R Bartoldson, Bhavya Kailkhura, Abhinav Bhatele, and Tom Goldstein.
\newblock Scaling up test-time compute with latent reasoning: A recurrent depth
  approach.
\newblock \emph{arXiv preprint arXiv:2502.05171}, 2025.

\bibitem[Hegazy et~al.(2026)Hegazy, Alanwar, and Elhoushi]{hegazy2026gated}
Amr Hegazy, Amr Alanwar, and Mostafa Elhoushi.
\newblock Gated recurrent transformers: Expressive depth through recurrent
  modulation in transformers.
\newblock \emph{arXiv preprint arXiv:2608.15062}, 2026.
\newblock \doi{10.48550/arXiv.2608.15062}.
\newblock URL \url{https://arxiv.org/abs/2608.15062}.

\bibitem[Jeddi et~al.(2026)Jeddi, Ciccone, and Taati]{jeddi2026loopformer}
Ahmadreza Jeddi, Marco Ciccone, and Babak Taati.
\newblock Loopformer: Elastic-depth looped transformers for latent reasoning
  via shortcut modulation.
\newblock \emph{arXiv preprint arXiv:2602.11451}, 2026.

\bibitem[McLeish et~al.(2025)McLeish, Li, Kirchenbauer, Kalra, Bartoldson,
  Kailkhura, Schwarzschild, Geiping, Goldstein, and
  Goldblum]{mcleish2025teaching}
Sean McLeish, Ang Li, John Kirchenbauer, Dayal~Singh Kalra, Brian~R Bartoldson,
  Bhavya Kailkhura, Avi Schwarzschild, Jonas Geiping, Tom Goldstein, and Micah
  Goldblum.
\newblock Teaching pretrained language models to think deeper with retrofitted
  recurrence.
\newblock \emph{arXiv preprint arXiv:2511.07384}, 2025.

\bibitem[Saunshi et~al.(2025)Saunshi, Dikkala, Li, Kumar, and
  Reddi]{saunshi2025reasoning}
Nikunj Saunshi, Nishanth Dikkala, Zhiyuan Li, Sanjiv Kumar, and Sashank~J
  Reddi.
\newblock Reasoning with latent thoughts: On the power of looped transformers.
\newblock \emph{arXiv preprint arXiv:2502.17416}, 2025.

\bibitem[Schwarzschild et~al.(2021)Schwarzschild, Borgnia, Gupta, Huang,
  Vishkin, Goldblum, and Goldstein]{schwarzschild2021can}
Avi Schwarzschild, Eitan Borgnia, Arjun Gupta, Furong Huang, Uzi Vishkin, Micah
  Goldblum, and Tom Goldstein.
\newblock Can you learn an algorithm? generalizing from easy to hard problems
  with recurrent networks.
\newblock \emph{Advances in Neural Information Processing Systems},
  34:\penalty0 6695--6706, 2021.

\bibitem[Wang et~al.(2025)Wang, Fu, Cai, Tang, Lyu, Fang, Zheng, Zhou, Zeng,
  Xiao, et~al.]{wang2025ultra}
Yudong Wang, Zixuan Fu, Jie Cai, Peijun Tang, Hongya Lyu, Yewei Fang, Zhi
  Zheng, Jie Zhou, Guoyang Zeng, Chaojun Xiao, et~al.
\newblock Ultra-fineweb: Efficient data filtering and verification for
  high-quality llm training data.
\newblock \emph{arXiv preprint arXiv:2505.05427}, 2025.

\bibitem[Wang et~al.(2026)Wang, Fu, Zhao, Zhao, Zhou, Lin, Lyu, Xue, Yi, Wang,
  et~al.]{wang2026data}
Yudong Wang, Zixuan Fu, Hengyu Zhao, Chen Zhao, Chuyue Zhou, Xinle Lin, Hongya
  Lyu, Shuaikang Xue, Yi~Yi, Yingjiao Wang, et~al.
\newblock Data science and technology towards agi part i: Tiered data
  management.
\newblock \emph{arXiv preprint arXiv:2602.09003}, 2026.

\bibitem[Zellers et~al.(2019)Zellers, Holtzman, Bisk, Farhadi, and
  Choi]{zellers2019hellaswag}
Rowan Zellers, Ari Holtzman, Yonatan Bisk, Ali Farhadi, and Yejin Choi.
\newblock Hellaswag: Can a machine really finish your sentence?
\newblock \emph{arXiv preprint arXiv:1905.07830}, 2019.

\bibitem[Zhu et~al.(2025)Zhu, Wang, Hua, Zhang, Li, Que, Wei, Wen, Yin, Xing,
  et~al.]{zhu2025scaling}
Rui-Jie Zhu, Zixuan Wang, Kai Hua, Tianyu Zhang, Ziniu Li, Haoran Que, Boyi
  Wei, Zixin Wen, Fan Yin, He~Xing, et~al.
\newblock Scaling latent reasoning via looped language models.
\newblock \emph{arXiv preprint arXiv:2510.25741}, 2025.

\end{thebibliography}
\bibliographystyle{plainnat}

\newpage

\appendix

\section{Synthetic Factual Recall}
\label{app:synthetic-task}

\subsection{Dataset and evaluation}
\label{app-subsec:dataset-details}

Our dataset is made up of fictional people, each with six attributes: birth city, birth day, birth month, university, major, and employer. Each attribute is selected from a fixed set of possible values, and one person's set of attributes is referred to as their biography. Together, the six attributes contain 33.39 bits of information per person, based on the total number of possible combinations. Names and attributes are encoded using several tokens, so the model cannot identify a person or answer from a single token.

When we train, we show each model five different permutations of each biography. Each permutation contains the same set of facts but presents them in different orders. When we evaluate these models, we show the model those same biography formats and check whether it predicts the missing value correctly.

We test each fact using five fixed biography templates, and a fact is only considered recalled if the model gives the exact answer on at least four of the templates. To calculate the recalled bits per person, we note that each recalled attribute contributes $\log_2(\text{number of possibilities})$ bits, which we then sum across all the attributes.

The runs use the same 50M-parameter model, data, optimizer, and four random seeds. We train separate runs with 15\%, 45\%, and 100\% of the total training FLOPs, and evaluate every model across serving depths \texttt{K=1,...,64}.

\subsection{Tokenized examples}

For example, consider one fictional person and two of their attributes:

\begin{table}[!htbp]
  \centering
  \caption{Example token sequences for one fictional person and two attributes. Names and attribute values use several tokens.}
  \label{tab:example-attributes}
  \begin{tabular}{ll}
    \toprule
    Part & Token sequence \\
    \midrule
    Person's name & \texttt{N17 N204 N6} \\
    Birth city & \texttt{V3 V11} \\
    Employer & \texttt{V6 V10} \\
    \bottomrule
  \end{tabular}
\end{table}

Then at training and evaluation the model would see the following:

\begin{table}[!htbp]
  \centering
  \caption{The same facts presented in different orders. Evaluation requires exact prediction of the two-token birth-city value.}
  \label{tab:example-orderings}
  \begin{tabular}{ll}
    \toprule
    Stage & What the model sees \\
    \midrule
    Training & \texttt{N17 N204 N6 [employer] V6 V10 [birth city] V3 V11} \\
    Evaluation & \texttt{N17 N204 N6 [birth city] \_\_\_ \_\_\_ [employer] V6 V10} \\
    Correct response & \texttt{V3 V11} \\
    \bottomrule
  \end{tabular}
\end{table}

\subsection{Fixed-depth specialization}

We see the same pattern when we train fixed-depth models at \texttt{K=1,2,4,8} and evaluate every model at every serving depth from \texttt{K=1} through \texttt{K=16}. Each model works best near the depth used during training, and recall drops sharply as the serving depth moves farther away.

\begin{figure}[!htbp]
  \centering
  \includegraphics[width=\linewidth]{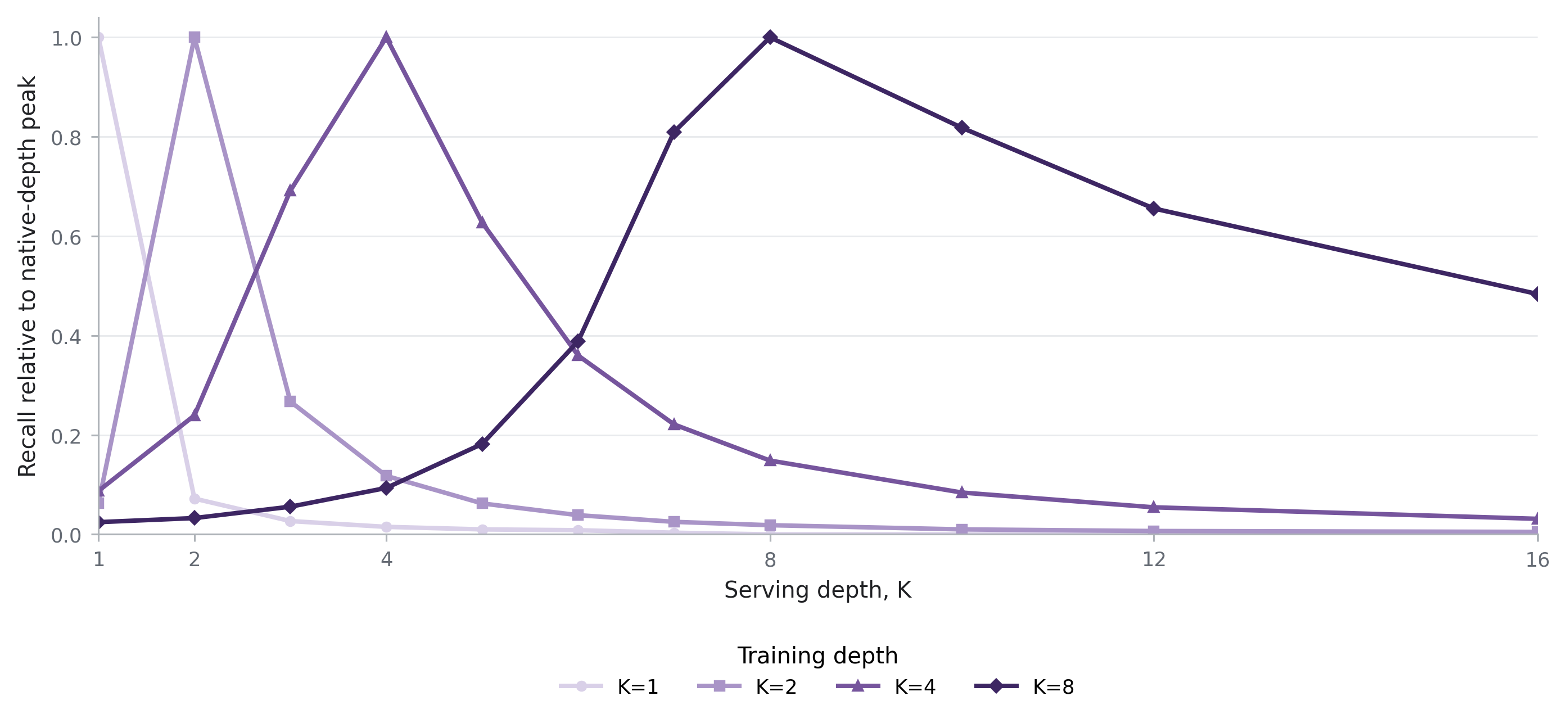}
  \caption{Recall across serving depths for fixed-depth models trained at \texttt{K=1}, \texttt{2}, \texttt{4}, or \texttt{8}, normalized by each model's peak recall. Each model performs best near the depth used during training.}
  \label{fig:fixed-depth-specialization}
\end{figure}

Fixed-depth training therefore only gives us half of what we want: it learns quickly, but it produces a depth-specialized model. This motivates using fixed depth for fast learning, then adding a short uniform phase to make what the model learned usable across the full serving range.

\FloatBarrier
\section{Experimental Setup}
\label{app:experimental-setup}

\subsection{Looped architecture and notation}

\begin{figure}[!htbp]
  \centering
  \includegraphics[width=\linewidth]{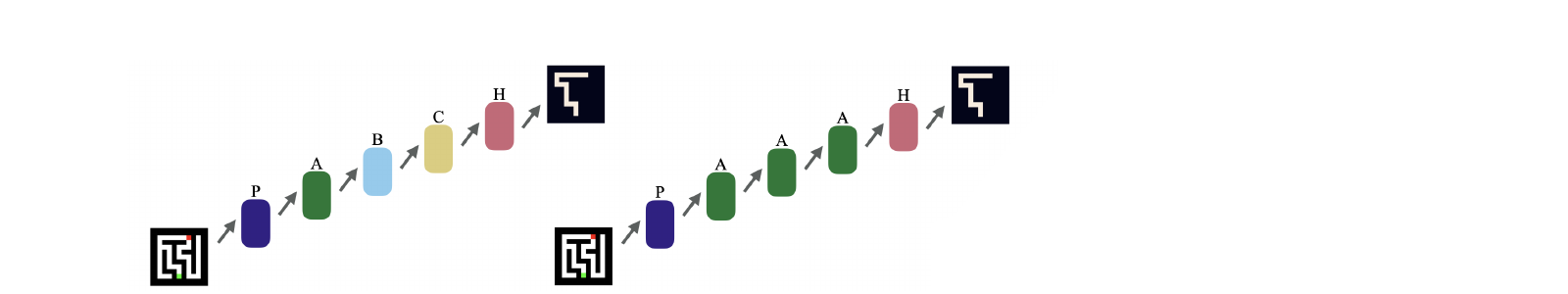}
  \caption{A standard transformer applies a sequence of different layers, while a looped transformer applies the same layer repeatedly. Here, the looped model applies layer ``A'' three times (\texttt{K=3}). ``\texttt{P}'' is the input embedding and ``\texttt{H}'' is the output head. Adapted, with permission, from \citet{schwarzschild2021can}.}
  \label{fig:architecture}
\end{figure}

\subsection{Depth schedules}

\begin{figure}[!htbp]
  \centering
  \includegraphics[width=\linewidth]{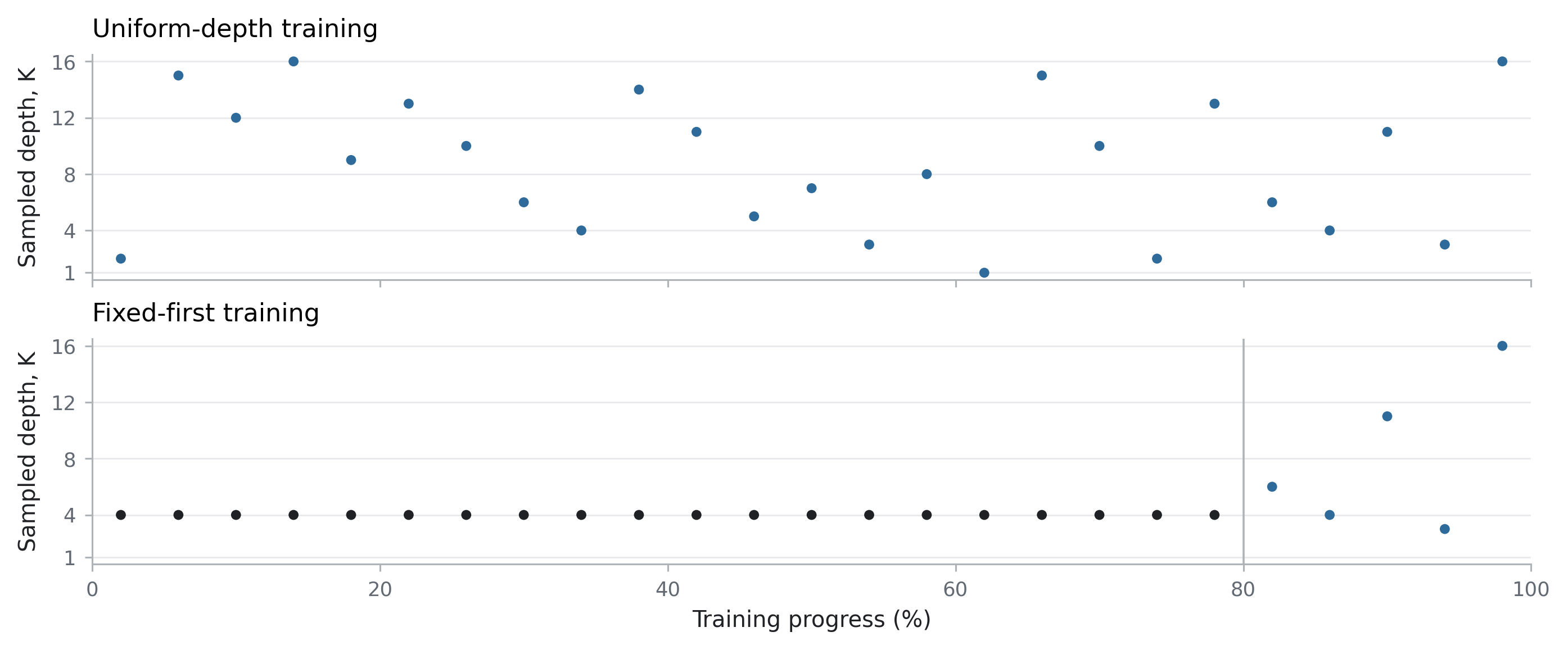}
  \caption{Uniform-depth training and an 80\% fixed-first schedule. Uniform-depth training samples \texttt{K} from 1 through 16 throughout training. Fixed-first training uses \texttt{K=4} for the first 80\% of training, then switches to the same uniform distribution.}
  \label{fig:schedule}
\end{figure}

Fixed-depth training uses the same value of \texttt{K} for every update. Uniform-depth training samples one value of \texttt{K} from a chosen serving range on every update, then computes the loss at that training depth. For the settings used here, uniform training has an average depth of 8.5, compared with 4 for fixed training.

The 200M sweep additionally uses $K_0\in\{1,2,4\}$ and either a uniform or Poisson-lognormal finishing distribution. The progressive \citet{mcleish2025teaching} reference uses their released 1-sqrt curriculum over the first 75\% of training, with the ramped mean set to \texttt{K=8} rather than 32 to match our \texttt{K<=16} serving range.

\subsection{Compute accounting and paired-run controls}

Modeled structural FLOPs are a hardware-independent estimate of the model computation implied by each sampled depth. Within each finishing family, paired runs use the same batches, optimizer settings, learning rate, token budget, and final depth sequence. Only the earlier depth schedule changes. Wall-clock reductions are measured against uniform training within each model scale.

\subsection{Language-model training and evaluation}

Both scales use a middle-loop decoder with two untied prelude blocks, an eight-block recurrent group applied \texttt{K} times, and two untied coda blocks. The 200M model has 193.5M parameters, hidden size 768, and 12 attention heads; the 1B model has 864.5M parameters, hidden size 2,048, and 32 attention heads.

The 200M models train on UltraData-Math-L3~\citep{wang2026data} for 4,800 updates, or 2.52B tokens. The 1B models train on Ultra-FineWeb~\citep{wang2025ultra} for 20,000 updates, or 10.49B tokens. Each schedule is run with three seeds, and every model is evaluated at serving depths \texttt{K=1,...,16}. The 1B evaluations are HellaSwag, PIQA, and ARC-Easy.

\FloatBarrier
\section{Complete Language-Model Results}
\label{app:language-model-results}

\subsection{Exact efficiency and terminal NLL}

\begin{table}[!htbp]
  \centering
  \caption{Cross-scale efficiency and NLL results. ``Fixed 67\%'' and ``Fixed 80\%'' use \texttt{K=4} for that portion of training, then switch to uniform \texttt{K=1,...,16}. Terminal NLL is measured over the final 200 updates, and 200M serving NLL is averaged over serving depths \texttt{K=1,...,16}. The compute-matched uniform run is a separate 3,296-update run at the same modeled FLOPs as Fixed 67\%, with its schedule re-tuned for the budget. Values are mean $\pm$ SD over three seeds.}
  \label{tab:cross-scale-results}
  \begingroup
  \small
  \setlength{\tabcolsep}{3pt}
  \begin{tabular}{lcccc}
    \toprule
    200M schedule & Terminal NLL & Mean serving NLL & Modeled FLOPs & Wall-clock reduction \\
    \midrule
    Uniform & 0.67297 $\pm$ 0.00058 & 0.68582 $\pm$ 0.00077 & 100\% & -- \\
    Fixed 67\% & 0.67424 $\pm$ 0.00027 & 0.68643 $\pm$ 0.00028 & 68.7\% & 26.2\% \\
    Uniform, matched & 0.71936 $\pm$ 0.00066 & 0.72089 $\pm$ 0.00072 & 68.7\% & -- \\
    Fixed 80\% & 0.67677 $\pm$ 0.00037 & 0.68941 $\pm$ 0.00087 & 62.4\% & 31.6\% \\
    \bottomrule
  \end{tabular}

  \vspace{0.6em}
  \begin{tabular}{lccc}
    \toprule
    1B schedule & Terminal NLL & Modeled FLOPs & Wall-clock reduction \\
    \midrule
    Uniform & 2.50349 $\pm$ 0.00177 & 100\% & -- \\
    Fixed 67\% & 2.51297 $\pm$ 0.00113 & 69.2\% & 32.1\% \\
    Fixed 80\% & 2.52264 $\pm$ 0.00195 & 63.1\% & 38.5\% \\
    \bottomrule
  \end{tabular}
  \endgroup
\end{table}

The loss jumps at the schedule transition because the objective changes from loss at \texttt{K=4} to loss sampled across \texttt{K=1} through \texttt{K=16}. Therefore, the values before and after the switch do not measure the same objective.

\FloatBarrier
\subsection{200M serving-depth NLL}

\begin{figure}[!htbp]
  \centering
  \includegraphics[width=\linewidth]{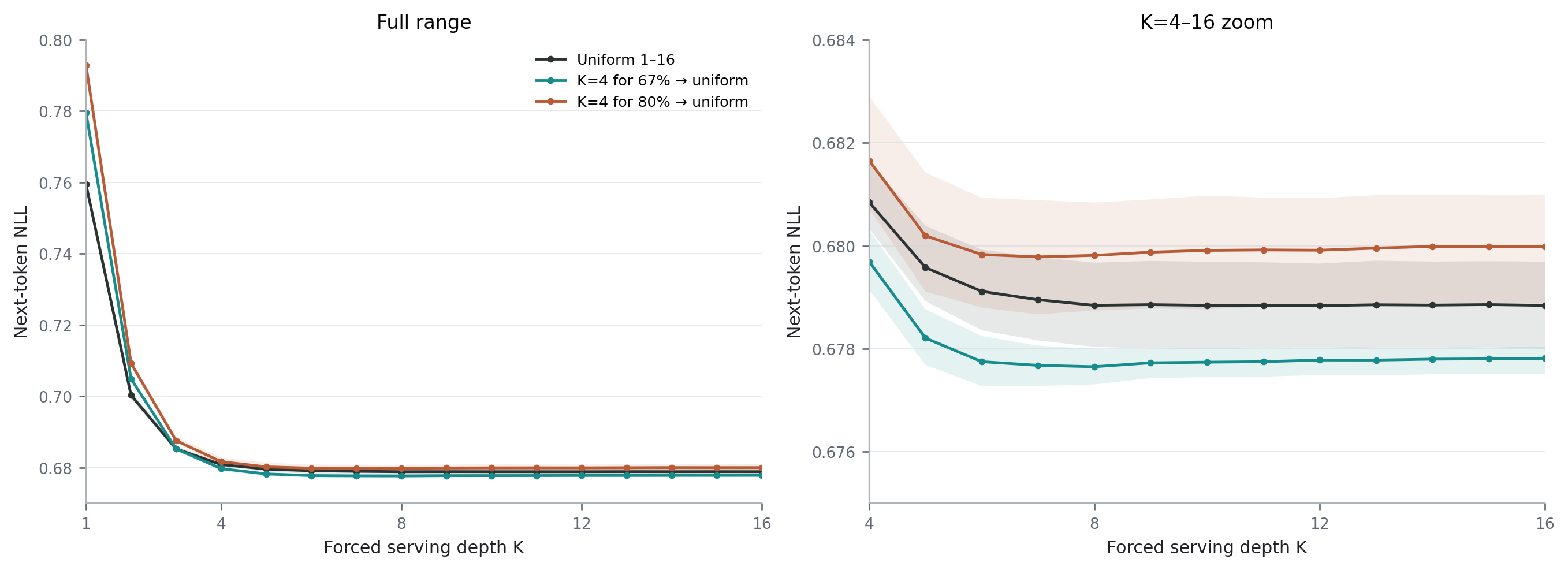}
  \caption{Next-token NLL across serving depths for the 200M models. The left panel shows the full serving range, and the right panel expands \texttt{K=4,...,16}. Both fixed-first schedules remain close to uniform training. Lines show three-seed means, and bands show the range across seeds.}
  \label{fig:small-serving-depth}
\end{figure}


\subsection{Exact 1B task results}

\begin{table}[!htbp]
  \centering
  \caption{Exact results for the three 1B evaluations shown in \Cref{fig:one-billion-results}. The top table averages each task separately over serving depths \texttt{K=1,...,16}; the bottom table reports the conventional \texttt{K=16} endpoint. There is no averaging across tasks. Values are mean $\pm$ SD over three seeds.}
  \label{tab:three-task-results}
  \begingroup
  \small
  \setlength{\tabcolsep}{3pt}
  \begin{tabular}{lccc}
    \toprule
    Depth-averaged task & Uniform & Fixed 67\% & Fixed 80\% \\
    \midrule
    HellaSwag & 50.72 $\pm$ 0.33 & 50.56 $\pm$ 0.19 & 50.27 $\pm$ 0.24 \\
    PIQA & 70.48 $\pm$ 0.42 & 70.52 $\pm$ 0.55 & 70.86 $\pm$ 0.53 \\
    ARC-Easy & 56.11 $\pm$ 1.46 & 56.76 $\pm$ 1.17 & 57.43 $\pm$ 1.12 \\
    \bottomrule
  \end{tabular}

  \vspace{0.7em}
  \begin{tabular}{lccc}
    \toprule
    Task at \texttt{K=16} & Uniform & Fixed 67\% & Fixed 80\% \\
    \midrule
    HellaSwag & 51.60 $\pm$ 0.41 & 51.51 $\pm$ 0.21 & 51.30 $\pm$ 0.30 \\
    PIQA & 70.73 $\pm$ 0.20 & 70.55 $\pm$ 0.57 & 70.91 $\pm$ 0.53 \\
    ARC-Easy & 56.20 $\pm$ 1.53 & 57.73 $\pm$ 0.97 & 57.95 $\pm$ 1.48 \\
    \bottomrule
  \end{tabular}
  \endgroup
\end{table}

\end{document}